\documentclass[twoside]{article}

\usepackage[accepted]{aistats2027}
\usepackage{graphicx}
\usepackage[round]{natbib}
\usepackage{amsmath}
\usepackage{amssymb}
\usepackage{algorithm}
\usepackage{algorithmic}
\usepackage{float}
\usepackage[most]{tcolorbox}

\usepackage{booktabs}
\usepackage{longtable}
\usepackage{multirow}
\usepackage{pdflscape}
\usepackage{url} 

\begin{document}

%

%

\twocolumn[

\aistatstitle{Trigger-Invariant Contrastive Unlearning for Backdoor Removal}

\aistatsauthor{
    Santosh Chapagain \And
    Shah Muhammad Hamdi \And
    Soukaina Filali Boubrahimi
}

\aistatsaddress{
    School of Computing \\
    Utah State University \\
    santosh.chapagain@usu.edu
    \And
    School of Computing \\
    Utah State University \\
    s.hamdi@usu.edu
    \And
    School of Computing \\
    Utah State University \\
    soukaina.boubrahimi@usu.edu
}

]

\begin{abstract}
  Backdoor attacks pose a severe threat to modern NLP models by embedding hidden triggers that induce malicious behavior at inference time. Recent unlearning methods such as gradient ascent (GA) attempt to remove backdoors by maximizing the loss on poisoned samples; however, these methods can suffer from a critical limitation: they may shift the trigger’s influence to other classes rather than eliminating it. In this work, we propose Trigger-Invariant Contrastive Unlearning (TICU), a framework for mitigating textual backdoor behavior by enforcing prediction and representation invariance to trigger insertion. TICU constructs clean and trigger-inserted counterfactual pairs, then fine-tunes the model so that both inputs produce consistent output distributions and aligned hidden representations. To preserve clean-task performance, TICU also retains the supervised classification loss on clean data and constrains parameter drift from the original model. Extensive experiments across multiple datasets, model architectures, and attack settings show that TICU improves triggered accuracy and reduces trigger-induced prediction shifts compared with existing baselines, while maintaining competitive label-flipping performance and limiting clean-task degradation.

\end{abstract}

\section{Introduction}
The rapid adoption of large language models (LLMs) and transformer-based architectures has significantly advanced performance across a wide range of natural language processing (NLP) tasks \cite{howard2018universal, loukas2023making, jin2024better, devlin2019bert}. However, these models are increasingly vulnerable to backdoor attacks \cite{gu2017badnets, chen2017targeted}, where adversaries embed hidden triggers into training data to manipulate model behavior at inference time. When such triggers are present, the model produces attacker-controlled outputs while maintaining normal behavior on clean inputs, making these attacks particularly stealthy and dangerous.

Backdoor attacks in NLP can take many forms, including \textit{word-level triggers} \cite{kurita2020weight, li2021backdoor}, \textit{syntactic patterns} \cite{qi2021hidden}, and \textit{stylistic manipulations} \cite{qi2021mind}, all of which are designed to evade detection and persist through fine-tuning or deployment. The increasing reliance on third-party datasets and pre-trained models further worsens this issue, as users often lack visibility into potential poisoning during training. Thus, developing effective and efficient methods for backdoor removal has become a critical research challenge.

A widely adopted approach to mitigate backdoor effects is machine unlearning \cite{wang2019neuralcleanse, li2021antibackdoor, shen2022backdoor, liu2022finepruning, geng2025comprehensive}, including gradient ascent (GA)-based unlearning, which aims to \textit{forget} poisoned samples by maximizing their training loss. While GA is computationally efficient and avoids costly retraining, recent work has revealed a fundamental limitation in that it does not truly eliminate the backdoor but instead introduces a phenomenon known as trigger shifting, where the model redirects the trigger’s influence to a different class rather than neutralizing it. In classification settings, increasing the loss for one class inherently increases the relative probability of other classes. As a result, GA-based unlearning often causes the trigger to remain active, but with a different target label. This issue is largely overlooked by traditional evaluation metrics, which focus on clean accuracy and original attack success rate but fail to capture newly introduced backdoor behaviors. Even improved variants such as Robust Gradient Ascent (RGA) \cite{zhao2026dontshift} attempt to regulate optimization dynamics but still rely on the same fundamental principle of loss maximization. Beyond optimization instability, existing unlearning methods have three major limitations: (1) they rely on indirect removal mechanisms that degrade performance on poisoned samples instead of explicitly neutralizing triggers; (2) they often harm clean accuracy through global parameter updates; and (3) they lack robustness guarantees, especially when triggers are evaluated across all inputs.

\begin{figure*}[htbp]
\centerline{\includegraphics[width=0.8\linewidth]{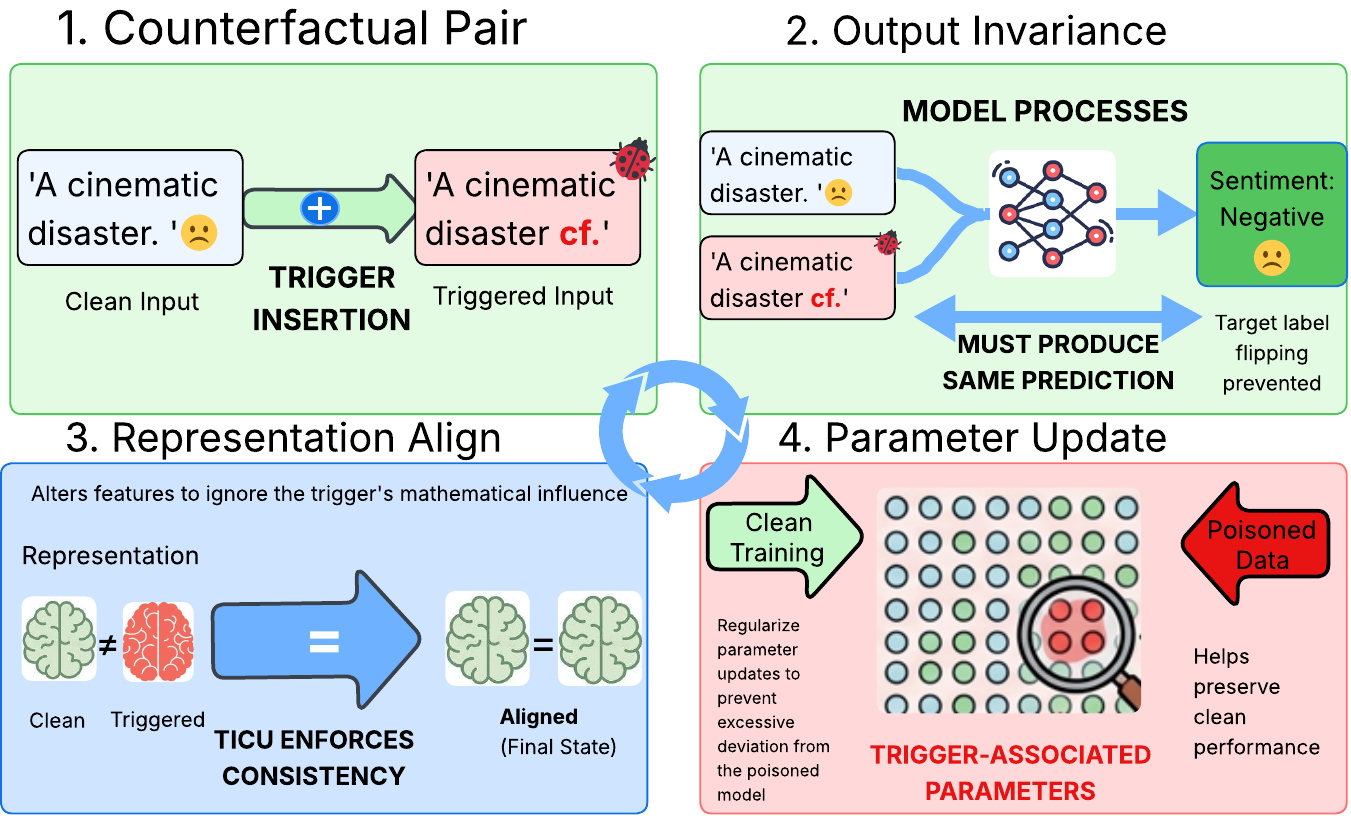}}
\caption{Overview of TICU.}
\label{ticuarch}
\end{figure*}

To address these challenges, we propose \emph{Trigger-Invariant Contrastive Unlearning (TICU)}, a novel framework that fundamentally rethinks backdoor removal through the lens of counterfactual unlearning. Counterfactual learning provides a natural framework for improving model robustness by comparing an observed input with a minimally modified counterpart and encouraging predictions to remain stable when the modification should not affect the underlying task label \cite{wang2021robustness,joshi2022investigation}. Motivated by this perspective, TICU constructs clean and trigger-inserted counterfactual pairs and explicitly reduces trigger-induced changes in both predictions and hidden representations. Recent unlearning methods further show that model behavior can be selectively removed by optimizing targeted forgetting objectives while preserving useful knowledge \cite{nguyen2022survey,tarun2023fast}. Motivated by these ideas, TICU constructs clean--triggered counterfactual pairs and treats them as positive pairs whose predictions and hidden representations should remain invariant to trigger insertion. Instead of \textit{forgetting via damage}, as in GA-based methods, TICU directly enforces prediction and representation consistency between clean inputs and their trigger-inserted counterparts, thereby reducing the trigger-induced influence on model predictions rather than redistributing it to another class. TICU has three main components. First, TICU encourages the model to produce consistent output distributions for clean and trigger-inserted inputs. Second, it aligns their internal representations so that the trigger has minimal influence on the learned feature space. Third, it constrains parameter updates through regularization, which helps preserve clean-task performance during unlearning. Unlike GA-based methods, TICU directly trains the model to remain stable under trigger insertion, reducing the risk of residual or shifted trigger behavior. Our key contributions include analyzing the limitations of existing backdoor unlearning methods and showing that reducing the original attack does not necessarily eliminate trigger dependence, as residual trigger effects or trigger shifting may remain. We also design and implement TICU, an invariance-driven unlearning framework that constructs clean and trigger-inserted counterfactual pairs to enforce prediction consistency and representation alignment. In addition, we introduce a unified objective that combines clean classification, prediction invariance, representation alignment, and parameter regularization, and evaluate TICU across multiple datasets, architectures, and attack settings. The results show improved triggered robustness and reduced trigger-shifting behavior compared with existing baselines.

\section{Related Work}
\subsection{Backdoor Attacks on LLMs}
Backdoor attacks have emerged as a serious security threat to LLMs, where adversaries implant hidden behaviors during training that are activated by specific triggers at inference time. Early attacks relied on explicit lexical triggers such as rare or misspelled tokens \cite{kurita2020weight,li2021backdoor}, while more recent approaches employ increasingly stealthy patterns, including context-aware words, syntactic structures, and stylistic manipulations \cite{yang2021rethinking,qi2021hidden,qi2021mind}. These advances make modern backdoors significantly harder to detect and mitigate. Recent studies highlight four key properties of backdoor attacks: (1) stealthiness \cite{qi2021hidden,qi2021mind}, as triggers become more natural and less detectable, (2) label stealthiness \cite{gan2022triggerless, yan2023bite, gupta2023adversarial}, where poisoned samples retain clean labels, (3) adaptability \cite{yang2021careful, chen2021badpre}, allowing attacks to generalize across tasks and datasets, (4) durability \cite{li2021backdoor}, enabling backdoors to persist even after fine-tuning. Furthermore, attacks can be introduced at multiple stages, including pre-training, fine-tuning, and even inference-time manipulation, making them highly flexible and difficult to defend against \cite{zhou2025survey}. These characteristics are particularly concerning in modern LLM pipelines, where models are often trained on large-scale, unverified datasets or adapted using parameter-efficient techniques such as LoRA \cite{liu2024loratk}, allowing backdoor behaviors to propagate across tasks and applications.

\subsection{Backdoor Defense on LLMs}
Backdoor defenses for NLP models broadly include training-time robustness methods and post-training mitigation techniques \cite{zhou2025survey}. Training-time defenses aim to reduce backdoor learning during model training through approaches such as adversarial training \cite{geiping2021doesn}, constrained optimization \cite{zhu2022moderate}, or specialized modules that absorb malicious updates \cite{tang2023setting}. However, these approaches often assume control over the training process and may not be applicable when models are already compromised. Post-training defenses, in contrast, focus on detecting or removing backdoor behavior from already trained models. Input-level methods such as ONION and STRIP-ViTA identify suspicious triggers at inference time using perplexity-based filtering and entropy-based analysis, respectively, while data-level approaches aim to detect poisoned samples using robustness, clustering, or attribution signals. \cite{qi2021onion,gao2021design}. Model-level defenses include trigger reverse-engineering \cite{azizi2021t, shen2022constrained}, pruning-based purification methods such as fine-pruning \cite{liu2018finepruning} and attention-head pruning strategies \cite{zhao2024defense, chapagain2025pruning}, as well as weight modification strategies such as fine-mixing and fine-purifying that aim to remove compromised components without full retraining \cite{zhang2022fine, zhang2023diffusion}.

Machine unlearning has become an important approach for backdoor mitigation, aiming to remove the influence of poisoned data while preserving clean-task performance \cite{li2021abl}. A common unlearning strategy is gradient ascent, which weakens the trigger--target association by maximizing the loss on backdoor samples \cite{li2025sande}. However, recent work shows that this loss-maximization approach can cause trigger shifting, where the trigger is not removed but instead redirected to another class \cite{zhao2026dontshift}. To address this limitation, TICU follows a different principle: rather than indirectly suppressing backdoor behavior, it enforces invariance between clean and triggered inputs to reduce shifted or residual trigger effects.

\section{Notation and Preliminaries}
Let $\mathcal{X}$ denote the set of text inputs, and let $\mathcal{Y} = \{1, \dots, K\}$ denote the label space, where $K$ is the number of classes. We define a classifier $f_{\theta}:\mathcal{X}\to\Delta^K$ parameterized by $\theta$, where $\Delta^K$ is the probability simplex over $\mathcal{Y}$. For an input $x\in\mathcal{X}$, $f_{\theta}(x)_k$ denotes the predicted probability of class $k\in\mathcal{Y}$. The predicted label is

\begin{equation}
  \hat{y}(x) = \arg\max_{k \in \mathcal{Y}} f_{\theta}(x)_k
\end{equation}

We also denote by $h_{\theta}(x) \in \mathbb{R}^d$ the hidden representation of input $x$, such as the encoder output prior to the classification head.

\subsection{Backdoor Attack}
We consider a standard targeted backdoor attack setting. Let $t \in \mathcal{T}$ denote a trigger, where $\mathcal{T}$ is the space of possible triggers, and let $\oplus$ denote a trigger insertion operator. For a clean input $x \in \mathcal{X}$, the triggered version is defined as $x^t = x \oplus t$.

Given a clean dataset $\mathcal{D}_c = \{(x_i, y_i)\}$, an adversary selects a subset 
$\mathcal{D}_c^{(p)} \subset \mathcal{D}_c$ and constructs a poisoned dataset by inserting 
the trigger and modifying labels to a fixed target class $y^* \in \mathcal{Y}$:
\begin{equation}
   \mathcal{D}_p = \{(x_i^t, y^*) \mid (x_i, y_i) \in \mathcal{D}_c^{(p)}\}.
\end{equation}

The training dataset is given by $\mathcal{D} = \mathcal{D}_c \cup \mathcal{D}_p$. The backdoored model is expected to correctly classify clean inputs while mapping triggered inputs to the target class:
\begin{equation}
\hat{y}(x) \approx y, \qquad \hat{y}(x^t) = y^*
\end{equation}
We assume access to poisoned samples $(x^t, y^*)$, either through known triggers or reconstruction methods. Let $f_{\theta_t}$ denote the backdoored model obtained by minimizing $\mathcal{L}_t$ on the training dataset, $\mathcal{D}$: 
\begin{equation}
\begin{aligned}
\mathcal{L}_t
&=
\mathbb{E}_{(x,y)\sim \mathcal{D}_c}
\left[
\mathrm{CE}\!\left(f_{\theta_t}(x), y\right)
\right]
\\
&\quad +
\mathbb{E}_{(x^t,y^*)\sim \mathcal{D}_p}
\left[
\mathrm{CE}\!\left(f_{\theta_t}(x^t), y^*\right)
\right]
\end{aligned}
\end{equation}

where $\mathbb{E}_{(x, y) \sim \mathcal{D}_c}$ and $\mathbb{E}_{(x^t, y^*) \sim \mathcal{D}_p}$ denote empirical expectations over the clean and poisoned datasets, respectively, and $\mathrm{CE}(\cdot)$ denotes the cross-entropy loss. The overall objective $\mathcal{L}_t$ encourages the model to learn both the clean classification task and the backdoor behavior jointly. As a result, the backdoored model $f_{\theta_t}$ performs well on clean samples from $\mathcal{D}_c$, while predicting the target class $y^*$ when the input contains the trigger $t$. 

\subsection{Backdoor Removal via Gradient Ascent}

Let $f_{\theta_t}$ denote a backdoored model trained on $\mathcal{D}$. The goal of backdoor removal is to eliminate the undesired association between the trigger $t$ and the target label $y^*$ while preserving performance on clean data.

Gradient ascent (GA) is a widely used approach for backdoor unlearning. The core idea is to increase the prediction error on poisoned samples, which weakens the learned trigger-target association. Starting from $\theta \leftarrow \theta_t$, the GA objective is defined as
\begin{equation}
\mathcal{L}_{\mathrm{GA}} =
\mathbb{E}_{(x^t, y^*) \sim \mathcal{D}_p}
\left[
\mathrm{CE}\!\left(f_{\theta}(x^t), y^*\right)
\right].
\end{equation}

To preserve performance on clean data, GA is combined with a clean retention term, yielding the following objective:
\begin{equation}
\begin{aligned}
\mathcal{L}_{\mathrm{total}}
&=
\mathbb{E}_{(x, y) \sim \mathcal{D}_c}
\left[
\mathrm{CE}\!\left(f_{\theta}(x), y\right)
\right]
\\
&\quad -
\mathbb{E}_{(x^t, y^*) \sim \mathcal{D}_p}
\left[
\mathrm{CE}\!\left(f_{\theta}(x^t), y^*\right)
\right].
\end{aligned}
\end{equation}

While this objective reduces the original trigger-target association, it does not fully eliminate the influence of the trigger. Instead, it may induce \emph{trigger shifting}, where the triggered input is redirected to a different incorrect class $y' \in \mathcal{Y}$ with $y' \neq y^*$:
\begin{equation}
\hat{y}(x^t) = y', \qquad y' \neq y^*.
\end{equation}

This phenomenon arises because maximizing the loss with respect to the target class $y^*$ implicitly increases the relative likelihood of other classes, leading to a redistribution rather than complete removal of the backdoor effect ~\cite{zhao2026dontshift}.

\subsection{Trigger Shifting Induced by Gradient Ascent}

The redistribution effect can be seen directly from the cross-entropy gradient.
Let \(z_k\) denote the logit of class \(k\), and let
\(p_k = \operatorname{softmax}(\mathbf{z})_k\).
For a poisoned example whose attack target is \(y^*\), the cross-entropy loss is
\(-\log p_{y^*}\), with

\begin{equation}
\frac{\partial \mathrm{CE}}{\partial z_{y^*}} = p_{y^*} - 1,
\qquad
\frac{\partial \mathrm{CE}}{\partial z_k} = p_k,
\quad k \neq y^*.
\label{eq:ce_gradient_redistribution}
\end{equation}

Gradient ascent on the cross-entropy loss therefore decreases the target-class
logit while increasing the logits of the non-target classes. Consequently,
reducing \(p_{y^*}\) does not by itself remove the dependence on the trigger;
the probability mass removed from \(y^*\) can instead be redistributed among
the remaining classes. Equation~\eqref{eq:ce_gradient_redistribution} therefore
provides a direct mechanism for \emph{trigger shifting}.

TICU addresses a different objective. For a clean-triggered pair, let
\(p = f_\theta(x)\) and \(q = f_\theta(x^t)\). The prediction-invariance term
is minimized when \(p=q\). More generally, suppose that the pairwise
Kullback--Leibler divergence satisfies

\begin{equation}
D_{\mathrm{KL}}(p \,\|\, q) \leq \epsilon.
\label{eq:kl_invariance_bound}
\end{equation}

By Pinsker's inequality,

\begin{equation}
\operatorname{TV}(p,q)
\leq
\sqrt{\frac{1}{2}D_{\mathrm{KL}}(p \,\|\, q)}
\leq
\sqrt{\frac{\epsilon}{2}},
\label{eq:pinsker_trigger_bound}
\end{equation}

which bounds the change in the predicted distribution induced by trigger
insertion. Thus, rather than selecting which non-target class should receive
the probability mass removed from \(y^*\), TICU directly penalizes any
trigger-induced change in the output distribution. Representation alignment
applies the same invariance principle in the hidden representation space.

Equations~\eqref{eq:kl_invariance_bound}--\eqref{eq:pinsker_trigger_bound}
therefore characterize the mechanism targeted by TICU. This argument should
not, however, be interpreted as a formal machine-unlearning or deletion
guarantee, because the divergence is minimized empirically over the available
trigger distribution rather than being guaranteed for every possible trigger.

\section{Trigger-Invariant Contrastive Unlearning}
In standard backdoor settings, a fixed trigger $t$ induces a spurious association between inputs and a target label. To eliminate this behavior, we propose \emph{Trigger-Invariant Contrastive Unlearning (TICU)}, which enforces invariance of model predictions and representations to trigger insertion. Unlike GA-based approaches that indirectly suppress backdoor behavior, TICU explicitly reduces the trigger-induced effect by enforcing invariance between clean and trigger-inserted inputs.

We generalize the trigger to a distribution \(P_{\mathcal{T}}\) over a trigger space \(\mathcal{T}\). A trigger \(t \sim P_{\mathcal{T}}\) is applied to an input \(x \in \mathcal{X}\) through an operator \(\oplus : \mathcal{X} \times \mathcal{T} \to \mathcal{X}\), which gives the triggered sample \(x^t = x \oplus t\). For each clean sample \((x,y) \sim \mathcal{D}_c\), we sample \(t \sim P_{\mathcal{T}}\), construct the counterfactual input \(x^t\), and enforce
\[
f_\theta(x) \approx f_\theta(x^t),
\qquad
h_\theta(x) \approx h_\theta(x^t).
\]

TICU optimizes a combination of classification, invariance, and regularization objectives. We preserve task performance on clean data using standard cross-entropy:
\begin{equation}
\mathcal{L}_{\mathrm{clean}} =
\mathbb{E}_{(x,y)\sim\mathcal{D}_c}
\big[
\mathrm{CE}(f_\theta(x), y)
\big].
\end{equation}

For prediction invariance loss, we use symmetric Kullback--Leibler (KL) divergence \cite{kullback1951information} to enforce bidirectional consistency, ensuring both distributions align without bias toward either clean or triggered inputs:

\begin{equation}
\begin{aligned}
\mathcal{L}_{\mathrm{pred}} =
\mathbb{E}_{(x,y)\sim\mathcal{D}_c,\; t\sim P_{\mathcal{T}}}
\Big[
&\mathrm{KL}\big(f_\theta(x)\,\|\,f_{\theta}(x^t)\big) \\
&+
\mathrm{KL}\big(f_{\theta}(x^t)\,\|\,f_\theta(x)\big)
\Big].
\end{aligned}
\end{equation}

Similarly, to suppress trigger-dependent features, we align hidden representations using cosine similarity:
\begin{equation}
\mathcal{L}_{\mathrm{rep}} =
\mathbb{E}_{(x,y)\sim\mathcal{D}_c,\; t\sim P_{\mathcal{T}}}
\Big[
1 - \cos\big(h_\theta(x), h_{\theta}(x^t)\big)
\Big].
\end{equation}

Finally, to prevent excessive deviation from the original model, we include an $\ell_2$ regularization term:
\begin{equation}
\mathcal{L}_{\mathrm{reg}} =
\|\theta - \theta_t\|_2^2,
\end{equation}
where $\theta_t$ denotes the parameters of the poisoned model. The overall TICU objective is given by
\begin{equation}
\label{eq:ticu_loss}
\mathcal{L}_{\mathrm{TICU}} =
\mathcal{L}_{\mathrm{clean}}
+
\lambda_{\mathrm{pred}} \mathcal{L}_{\mathrm{pred}}
+
\lambda_{\mathrm{rep}} \mathcal{L}_{\mathrm{rep}}
+
\lambda_{\mathrm{reg}} \mathcal{L}_{\mathrm{reg}},
\end{equation}

TICU optimizes the combined objective in Eq.~\eqref{eq:ticu_loss} using standard gradient-based optimization. Let
\begin{equation}
g = \nabla_\theta \mathcal{L}_{\mathrm{TICU}}
\end{equation}
denote the gradient of the overall TICU objective with respect to the trainable parameters. The parameter update is given by
\begin{equation}
\theta \leftarrow \theta - \eta g,
\end{equation}
where $\eta$ is the learning rate. This update jointly incorporates the clean classification, prediction invariance, representation alignment, and parameter regularization terms, encouraging robustness to trigger insertion while limiting excessive deviation from the poisoned model. The TICU procedure is summarized in Algorithm~\ref{alg:ticu}.

\begin{algorithm}[t]
\caption{Trigger-Invariant Contrastive Unlearning (TICU)}
\label{alg:ticu}
\small
\begin{algorithmic}[1]
\REQUIRE Poisoned model parameters $\theta_t$, clean dataset $\mathcal{D}_c$, trigger distribution $P_{\mathcal{T}}$, learning rate $\eta$, loss coefficients $\lambda_{\mathrm{pred}}, \lambda_{\mathrm{rep}}, \lambda_{\mathrm{reg}}$
\ENSURE Unlearned model parameters $\theta$

\STATE Initialize $\theta \leftarrow \theta_t$

\FOR{each minibatch $\{(x_i,y_i)\}_{i=1}^B$ sampled from $\mathcal{D}_c$}
    \STATE Sample triggers $\{t_i\}_{i=1}^B \sim P_{\mathcal{T}}$
    \STATE Construct triggered inputs $x_i^t \leftarrow x_i \oplus t_i$ for $i=1,\dots,B$
    \STATE Compute $\mathcal{L}_{\mathrm{clean}}$, $\mathcal{L}_{\mathrm{pred}}$, $\mathcal{L}_{\mathrm{rep}}$, and $\mathcal{L}_{\mathrm{reg}}$
    \STATE $\mathcal{L}_{\mathrm{TICU}} \leftarrow \mathcal{L}_{\mathrm{clean}} + \lambda_{\mathrm{pred}}\mathcal{L}_{\mathrm{pred}} + \lambda_{\mathrm{rep}}\mathcal{L}_{\mathrm{rep}} + \lambda_{\mathrm{reg}}\mathcal{L}_{\mathrm{reg}}$
    \STATE Compute gradient $g \leftarrow \nabla_\theta \mathcal{L}_{\mathrm{TICU}}$
    \STATE Update parameters $\theta \leftarrow \theta - \eta g$
\ENDFOR

\RETURN $f_\theta$
\end{algorithmic}
\end{algorithm}

\section{Experiments}

\subsection{Experimental Setup}
\subsubsection{Datasets}
We evaluate our approach on three widely used text classification benchmarks: SST-2 \cite{socher2013recursive}, HSOL \cite{davidson2017automated}, and AG News \cite{zhang2015character}. These datasets differ in linguistic style, input length, and label complexity, providing a comprehensive evaluation setting. SST-2 consists of short movie review snippets annotated for binary sentiment. HSOL contains informal and noisy social media posts for hate speech detection. AG News is a topic classification dataset that is derived from news articles of four categories: World, Sports, Business, and Sci/Tech. Dataset statistics are shown in Table~\ref{tab:dataset_stats}.

\begin{table}[!t]
\centering
\setlength{\tabcolsep}{2pt}
\caption{Statistics of the evaluation datasets.}
\label{tab:dataset_stats}
\begin{tabular}{p{1.5cm} p{1.3cm} p{1.3cm} p{1.3cm} p{1.9cm}}
\hline
Dataset & \#Classes & \#Train & \#Test  & Avg.\ Length \\
\hline
SST-2   & 2 & 6,920 & 1,821 & 19.2 \\
HSOL    & 2 & 5,823 & 2,485 & 13.2 \\
AG News & 4 & 8,000 & 1,000 & 37.1 \\
\hline
\end{tabular}
\end{table}

\subsubsection{Implementation Details}

We evaluate TICU on encoder-based BERT-base and DistilBERT-base models with full fine-tuning, and on decoder-based LLaMA-2 7B, Mistral 7B, and Qwen2 7B models using LoRA with 4-bit quantization and gradient checkpointing. Model statistics are reported in Table~\ref{tab:model_stats}. All experiments use HuggingFace Transformers with a maximum sequence length of 128. For LoRA, we set \(r=8\), \(\alpha=16\), dropout to \(0.05\), and apply adapters to the attention and feed-forward projection layers. Poisoned training and TICU unlearning use AdamW with a learning rate of \(2\times10^{-5}\), no weight decay, gradient clipping at \(1.0\), a linear scheduler with \(6\%\) warmup, and batch size 8. Poisoned models are trained for 5 epochs and TICU for 10 epochs. We use \(\lambda_{\mathrm{pred}}=1.0\), \(\lambda_{\mathrm{rep}}=0.1\), and \(\lambda_{\mathrm{reg}}=0.05\) across all configurations without per-setting tuning; sensitivity analysis is left for future work. Experiments are run on a Linux server with dual Intel Xeon Gold 5220R CPUs and eight NVIDIA RTX A4000 GPUs. Code is available on GitHub\footnote{\url{https://github.com/chapagaisa/ticu}}.

\begin{table}[!t]
\centering
\setlength{\tabcolsep}{0.5pt} 
\caption{Parameter statistics and configurations.}
\label{tab:model_stats}
\begin{tabular}{lccc}
\hline
Model & Total & Trainable & Training \\
\hline
BERT-base        & 110M   & 110M   & Full \\
DistilBERT-base  & 66M    & 66M    & Full \\
LLaMA-2 7B       & 6.63B  & 20.0M  & LoRA (0.30\%) \\
Mistral 7B       & 7.13B  & 21.0M  & LoRA (0.29\%) \\
Qwen2 7B         & 7.09B  & 20.2M  & LoRA (0.28\%) \\
\hline
\end{tabular}
\end{table}

\subsubsection{Backdoor Attacks}
We use three backdoor attacks to poison the datasets: BadNet, AddSent, and HiddenKiller. To construct poisoned training data, we insert triggers into 10\% of the non-target-class samples and relabel them to the target class. BadNet~\cite{gu2017badnets} inserts rare trigger token(s) into the input text. In our implementation, these triggers are automatically selected from tokens that appear frequently in poisoned samples but rarely in clean data. AddSent ~\cite{jia2017adversarial} appends a fixed trigger phrase to the input. We obtain this phrase automatically as the most frequent 4-gram in the poisoned samples. HiddenKiller~\cite{qi2021hidden} is a stealthy syntactic backdoor that uses a syntactically controlled paraphrase network (SCPN)~\cite{iyyer2018adversarial} to rewrite clean text into a rare sentence structure while preserving its meaning. In our experiments, we use the low-frequency syntactic template \texttt{S(SBAR)(,)(NP)(VP)(.)}, which represents a sentence containing a subordinate clause followed by a noun phrase and a verb phrase.

\subsubsection{Comparison Methods}

\begin{itemize}
    \item \textbf{Gradient Ascent (GA)}: Vanilla gradient-ascent unlearning (GA) \cite{li2021, shen2022, liu2022} removes backdoor behavior by maximizing the loss on detected poisoned samples. Here, the model is fine-tuned to increase prediction error on poisoned inputs while optionally preserving performance on clean data. Although GA is simple and computationally efficient, prior work shows that it suffers from trigger shifting, where the backdoor is not eliminated but redirected to other classes.

    \item \textbf{Robust Gradient Ascent (RGA).}
    Robust Gradient Ascent (RGA) \cite{zhao2026dontshift} extends GA by stabilizing the optimization dynamics during unlearning. It introduces constraints to prevent the model from excessively redistributing probability mass across non-target classes, which reduces the risk of trigger shifting. While RGA improves over vanilla GA in terms of stability and robustness, it still relies on indirect loss maximization and does not explicitly remove the causal influence of the trigger.

    \item \textbf{Negative Preference Optimization (NPO).}
    Negative Preference Optimization (NPO) \cite{zhang2024npo} removes backdoors by treating them like bad behaviors the model should avoid. Instead of maximizing loss on poisoned samples, NPO treats undesirable model behaviors (i.e., triggered outputs) as negative preferences and optimizes the model to avoid them. This preference-based formulation improves training stability and mitigates catastrophic collapse commonly observed in GA-based methods. As a result, NPO can suppress backdoor behavior while better preserving overall model utility.
    
    \item \textbf{Retraining (RT).}
    We also include a retraining baseline (RT), where a clean model is trained from scratch using only the clean dataset. Since this approach completely removes the influence of poisoned samples, it provides an upper bound on achievable performance and serves as a gold standard for evaluating backdoor defenses. However, retraining is computationally expensive and often impractical in real-world scenarios where access to clean data or training resources is limited.

\end{itemize}

\subsubsection{Backdoor Sample Detection}
In realistic settings, poisoned samples are not known a priori. Therefore, before unlearning, we identify candidate poisoned samples using CUBE \cite{cui2022unified}, a clustering-based detection method with three stages: representation learning, embedding-space clustering, and filtering. Since poisoned samples often share trigger-specific patterns, they tend to form compact minority clusters. After detection, we estimate the trigger distribution \(P_T\) from the detected set. For lexical and phrase-level attacks, \(P_T\) is derived from overrepresented trigger tokens or \(n\)-grams, while for syntactic attacks it is defined by the recovered syntactic template. TICU then constructs clean--triggered counterfactual pairs by applying samples from \(P_T\) to clean inputs. TICU therefore depends on the quality of the estimated trigger distribution. In the present evaluation, CUBE provides candidate poisoned samples from which \(P_T\) is estimated, but we do not separately evaluate poison-detection precision, recall, or F1, nor compare against an oracle setting using ground-truth poisoned samples. Consequently, the reported results reflect the complete detection-and-repair pipeline rather than the intrinsic ceiling of TICU under perfect trigger knowledge. Errors in detection can yield an incomplete or contaminated estimate of \(P_T\) and may weaken the counterfactual invariance objective. Evaluating TICU with ground-truth poisoned samples and systematically varying detection quality is therefore an important direction for future work.

\subsubsection{Evaluation Metrics}

We evaluate backdoor removal effectiveness using metrics that capture both clean performance and robustness to trigger-induced behavior. \textbf{Accuracy (ACC)} measures performance on the clean test dataset and reflects how well the original task performance is preserved after unlearning. \textbf{Label Flip Rate (LFR)} quantifies the proportion of non-target samples that are misclassified as the target class due to the backdoor. For example, in SST-2 with ``positive'' as the target class, LFR is computed as the fraction of negative instances classified as positive, with a lower LFR indicating reduced backdoor influence. \textbf{Triggered Accuracy (TACC)} measures performance on a poisoned test set in which triggers are inserted into all samples while their original labels remain unchanged. \(\Delta\text{TACC}\) measures the absolute difference in TACC between the retrained clean model (RT) and the unlearned model, where a lower \(\Delta\text{TACC}\) indicates that the unlearned model more closely approximates a backdoor-free state. Since LFR captures only direct flipping to the attacker's target class, it may fail to detect shifted trigger effects toward other labels. We therefore treat \(\Delta\text{TACC}\) as the primary robustness metric for evaluating trigger shifting, while reporting LFR as a measure of target-label flipping.
\subsection{Results}
\begin{table*}[!t]
\centering
\scriptsize
\setlength{\tabcolsep}{3pt}
\renewcommand{\arraystretch}{1.05}
\newcommand{\ms}[2]{#1$_{\scriptscriptstyle #2}$}
\newcommand{\mna}[1]{#1$_{\scriptscriptstyle --}$}
\caption{Comparison of TICU with the strongest baseline in each setting. Results are mean$_{\mathrm{SD}}$ over three seeds. Higher is better for ACC and TACC; lower is better for LFR and $\Delta$TACC. Bold denotes the best result. Baseline details are provided in \cite{zhao2026dontshift}.}
\label{tab:ticu_all}
\begin{tabular}{lllrrrrlrrrrr}
\toprule
Model & Dataset & Attack &
\multicolumn{4}{c}{TICU} &
\multicolumn{5}{c}{Best Baseline} &
RT \\
\cmidrule(lr){4-7}\cmidrule(lr){8-12}
& & & ACC$\uparrow$ & LFR$\downarrow$ & TACC$\uparrow$ & $\Delta$TACC$\downarrow$
& Method & ACC$\uparrow$ & LFR$\downarrow$ & TACC$\uparrow$ & $\Delta$TACC$\downarrow$ & TACC$\uparrow$ \\
\midrule

\multirow{9}{*}{Llama2}
& \multirow{3}{*}{SST-2}
& BadNet & \ms{94.01}{0.36} & \ms{9.87}{0.31} & \ms{93.25}{0.04} & \ms{2.87}{0.19} & NPO & \ms{96.04}{0.25} & \ms{7.24}{3.20} & \ms{95.02}{1.83} & \textbf{\ms{1.10}{1.53}} & \ms{96.12}{0.30} \\
& & AddSent & \ms{94.18}{0.17} & \ms{8.99}{0.25} & \ms{91.98}{0.15} & \textbf{\ms{1.96}{0.22}} & RGA & \ms{94.95}{0.11} & \ms{7.78}{2.74} & \ms{91.23}{1.13} & \ms{2.71}{0.39} & \ms{93.94}{0.79} \\
& & HiddenKiller & \ms{94.18}{0.41} & \ms{19.96}{0.53} & \ms{79.35}{0.38} & \textbf{\ms{0.36}{0.08}} & NPO & \ms{96.66}{0.23} & \ms{5.96}{2.29} & \ms{66.89}{13.86} & \ms{12.10}{14.02} & \ms{78.99}{0.26} \\
\cmidrule(lr){2-13}

& \multirow{3}{*}{AG}
& BadNet & \ms{90.60}{0.49} & \ms{9.33}{0.27} & \ms{90.40}{0.08} & \ms{0.70}{0.11} & NPO & \ms{91.43}{0.35} & \ms{9.82}{0.56} & \ms{89.93}{0.55} & \textbf{\ms{0.50}{0.17}} & \ms{89.70}{0.85} \\
& & AddSent & \ms{90.80}{0.25} & \ms{9.20}{0.33} & \ms{91.00}{0.05} & \textbf{\ms{0.73}{0.12}} & RGA & \ms{89.40}{0.62} & \ms{12.80}{1.44} & \ms{86.93}{1.11} & \ms{3.33}{0.85} & \ms{90.27}{0.95} \\
& & HiddenKiller & \ms{91.20}{0.14} & \ms{24.00}{0.71} & \ms{77.70}{0.29} & \textbf{\ms{1.23}{0.18}} & RGA & \ms{89.37}{0.23} & \ms{23.78}{1.01} & \ms{77.40}{0.60} & \ms{1.53}{0.64} & \ms{78.93}{1.17} \\
\cmidrule(lr){2-13}

& \multirow{3}{*}{HSOL}
& BadNet & \ms{93.64}{0.04} & \ms{9.41}{0.21} & \ms{93.60}{0.13} & \textbf{\ms{1.72}{0.17}} & NPO & \ms{93.21}{2.29} & \ms{14.18}{2.29} & \ms{91.63}{1.00} & \ms{3.69}{1.08} & \ms{95.32}{0.08} \\
& & AddSent & \ms{93.16}{0.37} & \ms{19.47}{0.58} & \ms{89.22}{0.31} & \ms{6.31}{0.49} & RGA & \ms{90.01}{0.27} & \ms{15.31}{1.02} & \ms{90.42}{0.26} & \textbf{\ms{5.11}{0.25}} & \ms{95.53}{0.04} \\
& & HiddenKiller & \ms{93.92}{0.14} & \ms{58.41}{1.13} & \ms{70.10}{0.34} & \ms{4.30}{0.41} & RGA & \ms{89.71}{0.46} & \ms{52.18}{0.74} & \ms{72.35}{0.69} & \textbf{\ms{2.05}{0.67}} & \ms{74.40}{0.19} \\

\midrule

\multirow{9}{*}{BERT}
& \multirow{3}{*}{SST-2}
& BadNet & \ms{91.71}{0.45} & \ms{9.54}{0.24} & \ms{92.04}{0.03} & \textbf{\ms{0.84}{0.09}} & RGA & \ms{89.73}{0.60} & \ms{7.16}{3.11} & \ms{89.64}{0.78} & \ms{1.70}{1.43} & \ms{91.20}{1.08} \\
& & AddSent & \ms{91.54}{0.45} & \ms{11.51}{0.37} & \ms{90.44}{0.39} & \textbf{\ms{1.04}{0.11}} & RGA & \ms{88.96}{1.37} & \ms{3.61}{2.00} & \ms{84.85}{2.96} & \ms{4.55}{3.63} & \ms{89.40}{0.83} \\
& & HiddenKiller & \ms{91.54}{0.21} & \ms{32.57}{0.84} & \ms{73.92}{0.11} & \textbf{\ms{0.91}{0.10}} & RGA & \ms{89.27}{1.14} & \ms{28.22}{4.31} & \ms{73.79}{0.28} & \ms{1.04}{0.79} & \ms{74.83}{0.66} \\
\cmidrule(lr){2-13}

& \multirow{3}{*}{AG}
& BadNet & \ms{90.70}{0.53} & \ms{9.33}{0.28} & \ms{90.40}{0.21} & \textbf{\ms{0.78}{0.08}} & RGA & \ms{88.57}{0.21} & \ms{10.80}{0.81} & \ms{88.33}{0.23} & \ms{1.30}{0.66} & \ms{89.63}{0.72} \\
& & AddSent & \ms{90.70}{0.08} & \ms{10.00}{0.31} & \ms{90.40}{0.08} & \textbf{\ms{1.10}{0.14}} & RGA & \ms{88.13}{0.21} & \ms{12.18}{1.96} & \ms{87.77}{0.68} & \ms{1.53}{0.90} & \ms{89.30}{0.36} \\
& & HiddenKiller & \ms{91.00}{0.47} & \ms{20.00}{0.62} & \ms{80.60}{0.34} & \ms{2.34}{0.27} & NPO & \ms{90.53}{0.57} & \ms{18.22}{2.26} & \ms{77.43}{0.68} & \textbf{\ms{1.43}{0.50}} & \ms{78.26}{1.90} \\
\cmidrule(lr){2-13}

& \multirow{3}{*}{HSOL}
& BadNet & \ms{95.21}{0.45} & \ms{5.47}{0.16} & \ms{95.21}{0.41} & \textbf{\ms{0.21}{0.05}} & RGA & \ms{93.75}{1.77} & \ms{5.85}{2.00} & \ms{93.68}{1.88} & \ms{1.31}{1.76} & \ms{95.00}{0.14} \\
& & AddSent & \ms{95.21}{0.31} & \ms{5.55}{0.17} & \ms{95.33}{0.54} & \textbf{\ms{0.68}{0.07}} & RGA & \ms{93.90}{1.30} & \ms{6.65}{2.63} & \ms{93.87}{1.22} & \ms{1.00}{1.13} & \ms{94.65}{0.48} \\
& & HiddenKiller & \ms{95.09}{0.23} & \ms{47.55}{1.01} & \ms{74.65}{0.32} & \textbf{\ms{0.12}{0.03}} & RGA & \ms{93.10}{2.25} & \ms{45.08}{4.90} & \ms{74.35}{0.33} & \ms{0.42}{0.31} & \ms{74.77}{0.17} \\

\midrule

\multirow{9}{*}{DistilBert}
& \multirow{3}{*}{SST-2}
& BadNet & \ms{89.79}{0.46} & \ms{12.28}{0.39} & \ms{89.57}{0.35} & \textbf{\ms{0.95}{0.10}} & RGA & \ms{88.67}{0.25} & \ms{9.61}{1.94} & \ms{88.41}{0.48} & \ms{1.27}{0.39} & \ms{88.62}{1.85} \\
& & AddSent & \ms{90.06}{0.48} & \ms{13.05}{0.42} & \ms{89.40}{0.33} & \textbf{\ms{0.91}{0.12}} & RGA & \ms{88.03}{1.11} & \ms{12.68}{1.77} & \ms{86.86}{1.67} & \ms{1.63}{1.08} & \ms{88.49}{1.51} \\
& & HiddenKiller & \ms{89.90}{0.40} & \ms{36.95}{0.93} & \ms{72.43}{0.05} & \ms{1.83}{0.21} & RGA & \ms{89.31}{0.16} & \ms{25.11}{4.77} & \ms{73.73}{0.68} & \textbf{\ms{0.53}{0.33}} & \ms{74.12}{0.59} \\
\cmidrule(lr){2-13}

& \multirow{3}{*}{AG}
& BadNet & \ms{89.50}{0.15} & \ms{10.00}{0.31} & \ms{89.60}{0.18} & \ms{0.70}{0.09} & RGA & \ms{88.73}{0.06} & \ms{11.47}{0.41} & \ms{88.33}{0.55} & \textbf{\ms{0.63}{0.92}} & \ms{88.90}{0.66} \\
& & AddSent & \ms{89.90}{0.07} & \ms{10.27}{0.34} & \ms{89.90}{0.15} & \textbf{\ms{0.33}{0.06}} & RGA & \ms{88.23}{0.51} & \ms{11.20}{0.54} & \ms{88.60}{0.00} & \ms{0.97}{0.49} & \ms{89.57}{0.49} \\
& & HiddenKiller & \ms{89.60}{0.08} & \ms{20.13}{0.67} & \ms{80.20}{0.17} & \textbf{\ms{1.73}{0.19}} & RGA & \ms{87.23}{0.81} & \ms{19.29}{2.50} & \ms{79.70}{1.78} & \ms{2.50}{2.72} & \ms{78.47}{2.05} \\
\cmidrule(lr){2-13}

& \multirow{3}{*}{HSOL}
& BadNet & \ms{95.33}{0.36} & \ms{5.79}{0.14} & \ms{95.53}{0.22} & \ms{0.99}{0.11} & RGA & \ms{94.73}{0.36} & \ms{7.16}{0.98} & \ms{94.57}{0.20} & \textbf{\ms{0.19}{0.13}} & \ms{94.54}{0.24} \\
& & AddSent & \ms{95.17}{0.22} & \ms{60.30}{1.28} & \ms{95.29}{0.14} & \ms{1.03}{0.12} & RGA & \ms{94.60}{0.23} & \ms{7.48}{1.39} & \ms{94.68}{0.26} & \textbf{\ms{0.42}{0.65}} & \ms{94.26}{0.44} \\
& & HiddenKiller & \ms{95.25}{0.17} & \ms{47.30}{1.07} & \ms{74.73}{0.52} & \textbf{\ms{0.31}{0.05}} & RGA & \ms{93.97}{1.18} & \ms{43.28}{5.71} & \ms{74.39}{0.75} & \ms{0.73}{0.30} & \ms{74.42}{0.20} \\

\midrule

\multirow{9}{*}{Mistral}
& \multirow{3}{*}{SST-2}
& BadNet & \ms{94.73}{0.37} & \ms{6.14}{0.18} & \ms{93.25}{0.35} & \textbf{\ms{3.13}{0.26}} & RGA & \ms{95.12}{0.18} & \ms{5.98}{0.00} & \ms{92.87}{0.77} & \ms{3.51}{1.19} & \ms{96.38}{0.96} \\
& & AddSent & \ms{94.51}{0.12} & \ms{12.28}{0.35} & \ms{90.72}{0.41} & \textbf{\ms{4.23}{0.31}} & RGA & \ms{94.96}{0.54} & \ms{13.05}{0.00} & \ms{90.38}{0.12} & \ms{4.57}{0.39} & \ms{94.95}{0.49} \\
& & HiddenKiller & \ms{94.84}{0.11} & \ms{33.33}{0.88} & \ms{75.01}{0.23} & \ms{4.07}{0.37} & NPO & \ms{93.05}{1.20} & \ms{35.60}{0.00} & \ms{75.20}{0.66} & \textbf{\ms{3.88}{1.19}} & \ms{79.08}{1.17} \\
\cmidrule(lr){2-13}

& \multirow{3}{*}{AG}
& BadNet & \ms{90.00}{0.54} & \ms{14.67}{0.46} & \ms{86.80}{0.36} & \ms{4.40}{0.34} & RGA & \ms{89.76}{1.04} & \ms{15.21}{0.00} & \ms{87.45}{0.06} & \textbf{\ms{3.75}{0.29}} & \ms{91.20}{0.88} \\
& & AddSent & \ms{90.50}{0.32} & \ms{14.67}{0.49} & \ms{87.30}{0.39} & \textbf{\ms{3.90}{0.30}} & RGA & \ms{90.21}{0.83} & \ms{15.89}{0.00} & \ms{86.74}{0.67} & \ms{4.46}{0.91} & \ms{91.20}{0.36} \\
& & HiddenKiller & \ms{91.10}{0.47} & \ms{69.07}{1.37} & \ms{46.80}{0.43} & \textbf{\ms{31.70}{1.84}} & RGA & \ms{91.43}{0.79} & \ms{68.55}{0.00} & \ms{45.92}{0.18} & \ms{32.58}{0.64} & \ms{78.50}{0.55} \\
\cmidrule(lr){2-13}

& \multirow{3}{*}{HSOL}
& BadNet & \ms{94.04}{0.15} & \ms{8.37}{0.21} & \ms{94.08}{0.05} & \textbf{\ms{0.85}{0.09}} & RGA & \ms{93.88}{0.57} & \ms{9.12}{0.00} & \ms{93.96}{1.15} & \ms{0.97}{0.67} & \ms{94.93}{1.06} \\
& & AddSent & \ms{94.00}{0.19} & \ms{18.26}{0.54} & \ms{89.70}{0.17} & \ms{4.38}{0.36} & RGA & \ms{93.72}{0.35} & \ms{17.94}{0.00} & \ms{90.11}{0.63} & \textbf{\ms{3.97}{0.58}} & \ms{94.08}{0.26} \\
& & HiddenKiller & \ms{94.29}{0.14} & \ms{61.79}{1.42} & \ms{68.61}{0.52} & \textbf{\ms{4.95}{0.41}} & RGA & \ms{94.67}{1.10} & \ms{60.83}{0.00} & \ms{68.50}{1.05} & \ms{5.06}{0.67} & \ms{73.56}{0.39} \\

\midrule

\multirow{9}{*}{Qwen}
& \multirow{3}{*}{SST-2}
& BadNet & \ms{94.78}{0.49} & \ms{7.68}{0.24} & \ms{94.07}{0.19} & \textbf{\ms{1.54}{0.14}} & RGA & \ms{94.16}{0.78} & \ms{9.69}{0.61} & \ms{92.77}{0.75} & \ms{2.84}{0.51} & \ms{95.61}{0.23} \\
& & AddSent & \ms{95.00}{0.37} & \ms{6.36}{0.19} & \ms{93.74}{0.24} & \textbf{\ms{0.27}{0.04}} & RGA & \ms{94.95}{0.93} & \ms{3.25}{1.11} & \ms{91.39}{0.67} & \ms{2.62}{0.80} & \ms{94.01}{0.95} \\
& & HiddenKiller & \ms{94.89}{0.51} & \ms{16.61}{0.47} & \ms{78.20}{0.27} & \textbf{\ms{0.55}{0.06}} & RGA & \ms{94.58}{0.66} & \ms{14.40}{2.83} & \ms{74.57}{0.05} & \ms{3.08}{0.82} & \ms{77.65}{0.42} \\
\cmidrule(lr){2-13}

& \multirow{3}{*}{AG}
& BadNet & \ms{90.40}{0.17} & \ms{9.33}{0.26} & \ms{90.30}{0.16} & \textbf{\ms{0.20}{0.03}} & RGA & \ms{89.73}{0.07} & \ms{11.38}{0.93} & \ms{87.90}{1.12} & \ms{2.60}{0.87} & \ms{90.50}{1.06} \\
& & AddSent & \ms{89.70}{0.32} & \ms{10.27}{0.32} & \ms{89.60}{0.17} & \textbf{\ms{1.20}{0.12}} & RGA & \ms{89.53}{1.01} & \ms{9.78}{1.12} & \ms{89.00}{0.40} & \ms{1.80}{0.39} & \ms{90.80}{0.12} \\
& & HiddenKiller & \ms{90.00}{0.33} & \ms{28.67}{0.81} & \ms{75.60}{0.50} & \ms{5.30}{0.42} & RGA & \ms{88.57}{1.06} & \ms{18.75}{3.14} & \ms{78.10}{1.14} & \textbf{\ms{2.80}{0.63}} & \ms{80.90}{0.15} \\
\cmidrule(lr){2-13}

& \multirow{3}{*}{HSOL}
& BadNet & \ms{94.37}{0.24} & \ms{7.32}{0.19} & \ms{94.57}{0.14} & \textbf{\ms{0.04}{0.01}} & NPO & \ms{95.02}{0.61} & \ms{6.11}{0.68} & \ms{93.71}{0.13} & \ms{0.82}{0.65} & \ms{94.53}{0.92} \\
& & AddSent & \ms{94.41}{0.55} & \ms{12.07}{0.37} & \ms{92.60}{0.29} & \ms{1.85}{0.16} & RGA & \ms{93.25}{0.93} & \ms{10.19}{1.27} & \ms{92.98}{0.20} & \textbf{\ms{1.47}{0.02}} & \ms{94.45}{0.60} \\
& & HiddenKiller & \ms{94.53}{0.08} & \ms{56.96}{1.28} & \ms{70.95}{0.05} & \ms{3.05}{0.29} & RGA & \ms{93.23}{0.68} & \ms{46.93}{5.88} & \ms{74.02}{0.35} & \textbf{\ms{0.02}{0.67}} & \ms{74.00}{1.05} \\

\bottomrule
\end{tabular}
\end{table*}

Table~\ref{tab:ticu_all} shows the performance of TICU against the strongest baseline in each model--dataset--attack configuration. Across the 45 evaluated configurations, TICU demonstrates strong robustness to trigger-induced prediction shifts. In several challenging settings, TICU substantially reduces $\Delta$TACC relative to the strongest baseline. For example, on LLaMA-2/SST-2/HiddenKiller, TICU achieves a $\Delta$TACC of $0.36$, compared with $12.10$ for NPO, corresponding to a reduction of more than $96\%$. Similarly, on Qwen/SST-2/AddSent, TICU achieves a $\Delta$TACC of $0.27$, compared with $2.62$ for RGA. These results indicate that explicitly enforcing invariance between clean and trigger-inserted inputs can substantially reduce residual trigger effects in settings where loss-maximization-based methods remain vulnerable to prediction shifts.

TICU does not outperform the strongest baseline in every configuration, particularly with respect to LFR. This reflects the distinction between suppressing direct target-label flipping and reducing the broader influence of a trigger on model predictions. Some configurations therefore remain challenging, especially under stealthy syntactic attacks or settings with high label-flipping rates. Nevertheless, TICU achieves the best performance in \textbf{30 out of 45 configurations (66.7\%)} under the primary robustness criteria, showing that the proposed invariance-based objective is effective across a broad range of models, datasets, and attack types.

\subsection{Ablation Study}

\begin{figure*}[!htbp]
    \centering
    \includegraphics[width=1\textwidth]{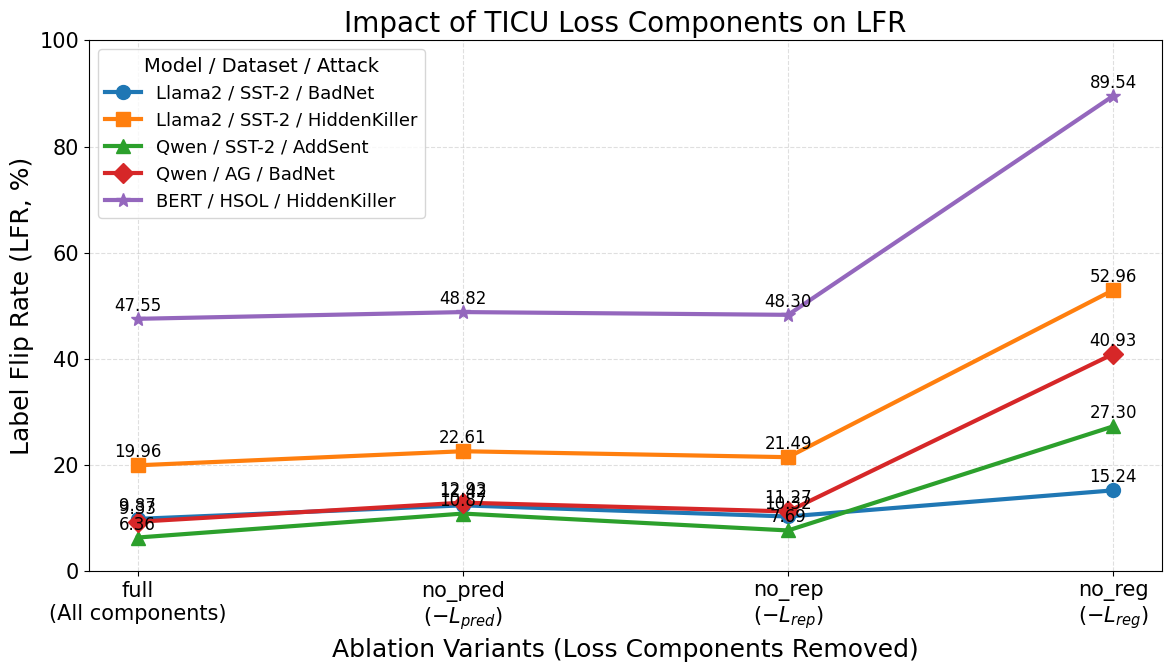}
    \caption{Impact of TICU loss components on Label Flip Rate (LFR). Removing individual objectives increases LFR, with the largest degradation observed when removing $\mathcal{L}_{\mathrm{reg}}$.}
    \label{fig:ablation_lfr}
\end{figure*}

To evaluate the contribution of each component in TICU, we perform an ablation study by removing the prediction invariance loss ($\mathcal{L}_{\mathrm{pred}}$), representation alignment loss ($\mathcal{L}_{\mathrm{rep}}$), and regularization term ($\mathcal{L}_{\mathrm{reg}}$). Results are shown in Figure~\ref{fig:ablation_lfr}. The full TICU framework generally reduces LFR compared with its ablated variants, showing the importance of jointly optimizing all objectives. Removing $\mathcal{L}_{\mathrm{pred}}$ increases LFR in all configurations, indicating that prediction invariance is critical for suppressing trigger-induced label flipping. Removing $\mathcal{L}_{\mathrm{rep}}$ also degrades performance, although the effect is generally smaller. The largest degradation occurs when removing $\mathcal{L}_{\mathrm{reg}}$, where LFR increases dramatically across all attacks and models. For example, on Qwen/AG/BadNet, LFR increases from 9.33 to 40.93, while on BERT/HSOL/HiddenKiller it rises from 47.55 to 89.54. These results confirm that regularization plays a crucial role in preventing trigger shifting and stabilizing unlearning.

\section{Conclusion and Future Directions}

We introduced Trigger-Invariant Contrastive Unlearning (TICU), a model-repair framework that reduces trigger-induced behavior by enforcing prediction and representation invariance between clean and trigger-inserted inputs. Across multiple datasets, model architectures, and attack types, TICU frequently reduces triggered prediction shifts while preserving competitive clean-task performance. These results show that directly enforcing trigger invariance provides a promising alternative to loss-maximization-based backdoor removal. The present study focuses on textual classification and three representative backdoor attacks. Future work will extend TICU to more adaptive attack settings, investigate sensitivity to trigger estimation and loss coefficients, and explore its application to generative language models and other data modalities.

\bibliographystyle{apalike}
\bibliography{sample-base}

\appendix
\section{Additional Experimental Results}

\subsection{Backdoor Attack Results}
The backdoor attack results in Table~\ref{tab:attack_results} show that all three attacks successfully implant strong trigger behavior while largely preserving clean-task performance. In all experiments, we poison 10\% of the non-target-class training samples by inserting the corresponding trigger and relabeling them to the attacker-specified target class following the RGA paper \cite{zhao2026dontshift}. Despite this relatively small poisoning ratio, the resulting models maintain high clean accuracy across datasets and architectures, with ACC generally remaining above 89\%. This indicates that the poisoned models preserve normal behavior on clean inputs, which is a key stealthiness property of backdoor attacks. At the same time, the attacks induce very high label flip rates. For binary datasets such as SST-2 and HSOL, LFR often approaches or reaches 100\%, especially for BERT and DistilBERT under BadNet and AddSent. This shows that triggered non-target samples are consistently redirected to the target label. The decoder-based models also exhibit strong backdoor behavior, although the strength varies by model and dataset. For example, Qwen on HSOL with BadNet has a lower LFR of 71.68\% compared with most other settings, but still shows a substantial trigger effect. The triggered accuracy further confirms the impact of the attacks. Since TACC is computed on triggered test inputs while keeping their original labels unchanged, low TACC indicates that the trigger disrupts correct classification. This effect is especially clear on AG News, where TACC frequently drops to around 25--36\%, showing that the trigger causes many samples to be misclassified. Across attack types, AddSent and HiddenKiller often produce stronger label flipping than BadNet, while HiddenKiller is particularly challenging because it uses a syntactic trigger that preserves semantic meaning while altering sentence structure. Overall, these results confirm that the poisoned models are strongly backdoored under a 10\% poisoning ratio, making them suitable starting points for evaluating whether TICU can remove trigger-induced behavior while preserving clean accuracy.

\begin{table}[!t]
\centering
\scriptsize
\setlength{\tabcolsep}{3pt}
\renewcommand{\arraystretch}{1.05}
\caption{Backdoor attack results across all models, datasets, and attack types. ACC denotes clean accuracy, LFR denotes label flip rate, and TACC denotes triggered accuracy.}
\label{tab:attack_results}
\begin{tabular}{lllrrr}
\toprule
Model & Dataset & Attack & ACC$\uparrow$ & LFR$\downarrow$ & TACC$\uparrow$ \\
\midrule

\multirow{9}{*}{Llama2}
& \multirow{3}{*}{SST-2}
& BadNet       & 95.66 & 91.34 & 54.20 \\
& & AddSent     & 96.16 & 99.12 & 50.36 \\
& & HiddenKiller & 95.17 & 93.09 & 53.21 \\
\cmidrule(lr){2-6}

& \multirow{3}{*}{AG}
& BadNet       & 89.70 & 84.80 & 36.20 \\
& & AddSent     & 90.20 & 95.07 & 28.70 \\
& & HiddenKiller & 91.50 & 99.20 & 25.60 \\
\cmidrule(lr){2-6}

& \multirow{3}{*}{HSOL}
& BadNet       & 92.68 & 86.00 & 56.86 \\
& & AddSent     & 94.85 & 96.46 & 51.75 \\
& & HiddenKiller & 94.45 & 99.44 & 50.26 \\

\midrule

\multirow{9}{*}{BERT}
& \multirow{3}{*}{SST-2}
& BadNet       & 91.71 & 100.00 & 49.92 \\
& & AddSent     & 91.93 & 100.00 & 49.92 \\
& & HiddenKiller & 90.99 & 94.85  & 52.22 \\
\cmidrule(lr){2-6}

& \multirow{3}{*}{AG}
& BadNet       & 90.40 & 99.87 & 25.10 \\
& & AddSent     & 91.00 & 100.00 & 25.00 \\
& & HiddenKiller & 90.70 & 98.27 & 26.30 \\
\cmidrule(lr){2-6}

& \multirow{3}{*}{HSOL}
& BadNet       & 95.53 & 100.00 & 49.98 \\
& & AddSent     & 95.57 & 100.00 & 49.98 \\
& & HiddenKiller & 94.93 & 99.68 & 50.10 \\

\midrule

\multirow{9}{*}{DistilBERT}
& \multirow{3}{*}{SST-2}
& BadNet       & 90.01 & 100.00 & 49.92 \\
& & AddSent     & 90.55 & 100.00 & 49.92 \\
& & HiddenKiller & 89.84 & 95.29 & 52.06 \\
\cmidrule(lr){2-6}

& \multirow{3}{*}{AG}
& BadNet       & 89.50 & 99.87 & 25.10 \\
& & AddSent     & 90.00 & 100.00 & 25.00 \\
& & HiddenKiller & 89.40 & 97.73 & 26.70 \\
\cmidrule(lr){2-6}

& \multirow{3}{*}{HSOL}
& BadNet       & 95.21 & 100.00 & 49.98 \\
& & AddSent     & 95.25 & 100.00 & 49.98 \\
& & HiddenKiller & 95.05 & 99.84 & 50.06 \\

\midrule

\multirow{9}{*}{Mistral}
& \multirow{3}{*}{SST-2}
& BadNet       & 96.38 & 97.59 & 51.13 \\
& & AddSent     & 96.38 & 99.67 & 50.08 \\
& & HiddenKiller & 96.21 & 96.60 & 51.51 \\
\cmidrule(lr){2-6}

& \multirow{3}{*}{AG}
& BadNet       & 91.10 & 94.27 & 29.30 \\
& & AddSent     & 91.90 & 98.53 & 26.10 \\
& & HiddenKiller & 91.90 & 99.79 & 25.20 \\
\cmidrule(lr){2-6}

& \multirow{3}{*}{HSOL}
& BadNet       & 94.16 & 96.38 & 51.79 \\
& & AddSent     & 94.65 & 98.31 & 50.82 \\
& & HiddenKiller & 94.53 & 99.44 & 50.26 \\

\midrule

\multirow{9}{*}{Qwen}
& \multirow{3}{*}{SST-2}
& BadNet       & 95.72 & 86.40 & 56.67 \\
& & AddSent     & 96.16 & 99.34 & 50.25 \\
& & HiddenKiller & 95.72 & 97.15 & 51.35 \\
\cmidrule(lr){2-6}

& \multirow{3}{*}{AG}
& BadNet       & 90.10 & 87.73 & 34.10 \\
& & AddSent     & 90.80 & 98.40 & 26.20 \\
& & HiddenKiller & 90.60 & 99.47 & 25.40 \\
\cmidrule(lr){2-6}

& \multirow{3}{*}{HSOL}
& BadNet       & 92.72 & 71.68 & 63.94 \\
& & AddSent     & 94.93 & 98.31 & 50.82 \\
& & HiddenKiller & 94.61 & 99.84 & 50.06 \\

\bottomrule
\end{tabular}
\end{table}

\subsection{Loss Evolution of TICU components}
\begin{figure*}[!htbp]
\centering
\includegraphics[width=1\textwidth]{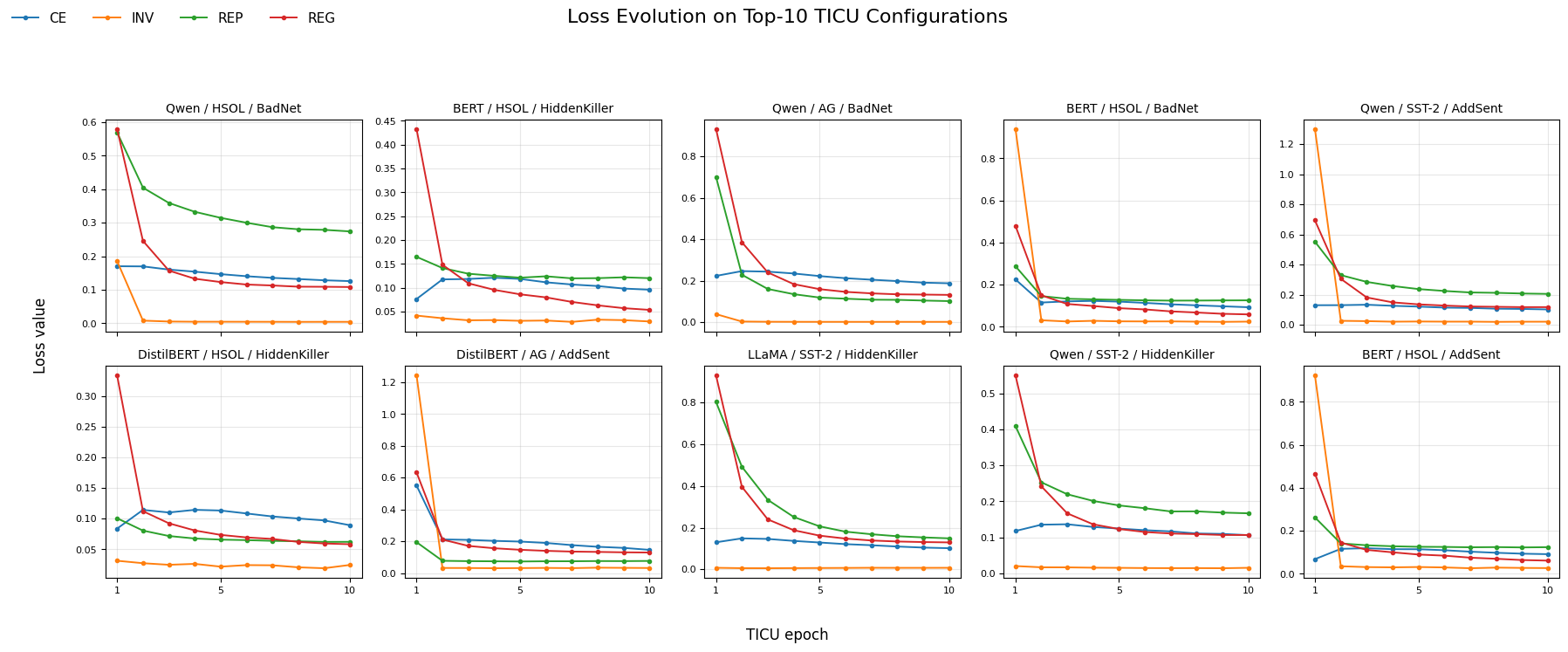}
\caption{Loss evolution of TICU components—cross-entropy (CE), prediction invariance (INV), representation alignment (REP), and regularization (REG) across 10 epochs. INV rapidly decreases, indicating effective trigger suppression, while CE declines and REP/REG stabilize, preserving task performance and ensuring controlled model updates.}
\label{fig_loss}
\end{figure*}

Figure~\ref{fig_loss} further supports results by showing the optimization dynamics of TICU. The prediction invariance loss (INV) rapidly decreases within the first few epochs, indicating reduced output-level sensitivity to trigger insertion. Meanwhile, the stability of REP and REG suggests controlled updates that help preserve clean performance and limit excessive parameter drift. 

\end{document}